# Can LLMs Really Understand Item Difficulty Levels? Implications for Automated Item Generation Using LLMs

Xinyi Wang[1,2], Hong Jiao[1], Ming Li[1], Sydney Peters[1], Hanna Choi[1],

Tianyi Zhou[3], Qingshu Xu[4]

[1]University of Maryland, College Park

[2]Beijing Normal University, China

[3]Mohamed bin Zayed University of Artificial Intelligence

[4]Shandong Normal University

**Abstract**

The estimation of item difficulty plays a key role in both formative assessment and large-scale high-stakes summative assessments. This study explores how large language models (LLMs) performed in predicting item difficulty levels using items from a large-scale Reading and Writing test. The study investigated various prompting strategies and parameter settings across multiple LLMs. LLM performance was compared with encoder-only language models and feature-based supervised machine learning models. Zero-shot GPT-4.1 with a temperature of 0 yielded the highest item difficulty level prediction accuracy with a quadratic weighted kappa (QWK) of 0.578. However, LLMs' prediction accuracy was lower than ConvBERT (QWK=0.625), which outperformed the best feature-based supervised machine learning model. Further analysis showed that all LLMs struggled to label hard items, particularly the current advanced GPT 5.4 tended to underestimate item difficulty levels. Dimension reduction of embedding showed that item embedding from different difficulty levels were mixed together, indicating that sematic information from items alone is likely insufficient for item difficulty level prediction. The implications of the findings highlighted that if LLMs cannot understand item difficulty levels as evidenced by empirical data and treated most items as easy items when their own capabilities increased, caution should be exercised in generating items with targeted item difficulty levels.

## 1. Introduction

Estimation of item difficulty plays a key role for both formative assessment and large-scale high-stakes summative assessments (AlKhuzaey et al., 2024; Li, Jiao, et al., 2025; Peters et al., 2025). First, item difficulty is an important indicator of item quality. When an item is too easy or too difficult, less information could be provided to distinguish test-takers with different ability levels. Further, the adequate spread and the adequate number of items at different difficulty/ability levels assures an item bank's width and depth to provide high measurement precision at different ability levels, especially a requirement for high-quality implementation of computerized adaptive test (CAT) based on item response theory (IRT) models (Wauters et al., 2012). In addition, the information about the difficulty levels of newly developed items before item piloting helps to construct parallel operational test forms which consist of both core operational items (for scoring) and field-test items (for new item piloting) for large-scale assessments (Lane et al., 2016).

The common practice to estimate item difficulty is to collect item response data through field-testing, which can be time-consuming and costly (AlKhuzaey et al., 2024; Hsu et al., 2018). Since new items are often added to field-test forms without prior difficulty information, field-test forms may differ in difficulty and lead to unequal test-taking experiences. Item statistics are then reviewed to determine whether newly piloted items would be accepted, revised, or rejected. This item quality review process may take several months or ever a year. Thus, such practices may not help efficient replenishing of an item bank with targeted ranges of item difficulty.

Text-based approaches to predicting item difficulty have been explored as an alternative to item piloting (Yancey et al., 2024), and some studies have shown promising results (Ha et al., 2019; Li, Jiao, et al., 2025; Peters et al., 2025). Linguistic and cognitive features of items are extracted and used in supervised machine learning models for item difficulty prediction. Large language models (LLMs), trained on massive text corpora with emerging reasoning capabilities, may provide a promising approach to item difficulty prediction. With the recent development of LLMs, several studies (Benedetto et al., 2024; Li, Jiao, et al., 2025; Park et al., 2024; Zotos et al., 2025) evaluate the performance of LLMs in understanding and predicting item difficulty. However, none study has explored LLMs' capabilities in predicting item difficulty levels. This study investigated the capabilities of LLMs in item difficulty level prediction. Several factors were examined including different LLMs, prompting strategies, and LLM parameter settings. Their performance was further compared with that of linguistic feature-based supervised machine learning models and encoder-only language models.

## 2. Related Works

Text-based approaches to predicting item difficulty have been explored for decades (AlKhuzaey et al., 2024; Benedetto, 2023; Ferrara et al., 2022a; Peters et al., 2025). In the early stages, the prediction models mainly relied on handcrafted surface-level textual features. As an early example of item difficulty modeling, Drum et al. (1981) used surface structure variables from passages, question stems, correct answers, and distractors to predict item p-values in standardized reading comprehension tests. Their results showed that these surface structure variables accounted for approximately three-quarters of the variation in item p-values. Perkins et al. (1995) applied artificial neural networks (ANNs) to predict item difficulty in a standardized reading comprehension test. Using text structure (e.g., the number of words), propositional analysis of passages and stems (e.g. the number of arguments), and cognitive demand, their ANN model predicted item difficulty (p-values), and achieved a mean squared error (MSE) of 0.0129. With advances in natural language processing (NLP), researchers started to use NLP tools to automatically extract more linguistic features from item texts, such as readability indices (Benedetto et al., 2020; Ferrara et al., 2022b; Štěpánek et al., 2023). Yi et al. (2024) proposed an interpretable machine learning method for predicting mathematics item difficulty using data from a college entrance examination. They classified items into three difficulty levels: easy, moderate and difficult. Their model used eight features, including reasoning steps, calculations complexity, knowledge content and cognitive level, as predictors in an XGBoost classifier with grid-search hyperparameter tuning, achieving an accuracy of 0.806.

As NLP techniques advanced, both static and contextual embeddings were developed. Studies (AlKhuzaey et al., 2024; Hsu et al., 2018; Peters et al., 2025) explored embedding-based representations, which can capture deeper semantic information contained in items. Early word embedding models, such as Word2Vec and GloVe, assign each word a fixed vector representation. Later models, such as ELMo introduce contextual embeddings, allowing word representations to vary according to context. Hsu et al. (2018) used Word2Vec to transform item texts into semantic vectors and then extracted cosine-similarity features to develop a support vector machine (SVM) classifier. Their results showed that semantic stem-answer similarity was the strongest predictor. Ha et al. (2019) used linguistic features, information retrieval features and embedding-based representations extracted with Word2Vec and ELMo to predict the difficulty of multiple-choice (MC) items from high-stake medical exams and showed embeddings had the strongest predictive power.

More recently, encoder-only transformer models, such as BERT (Devlin et al., 2019) and its variants, improved contextual representation, capturing nuanced semantic and contextual relationship. Zhou and Tao (2020) pre-trained BERT on a task-specific corpus for predicting the difficulty of programming problems and achieved an accuracy of 0.67 for binary item difficulty level prediction, outperforming basic BERT and BiLSTM. Bulut et al. (2024) extracted Coh-Metrix linguistic features, BiomedBERT embeddings, and stem-option semantic similarity features from stems and options of MC items for the United States Medical Licensing Examination (USMLE), applied Principal Component Analysis (PCA) for dimensionality reduction of features, and lasso regression achieved a root mean square error (RMSE) of 0.253 for transformed p-value.

The emergence of LLMs introduced a new direction for item difficulty prediction. LLMs are trained on broad and diverse corpora, and can capture contextual meaning, domain knowledge and even reasoning patterns. Recent studies have explored the use of LLMs for predicting item difficulty, as student response simulators, feature extractors, and evaluators of the cognitive demands of items. Benedetto et al. (2024) explored whether LLMs (GPT-3.5-

turbo-0613, GPT-3.5-turbo-1106, and GPT-4-1104-preview) can be used to simulate students' responses to three public MC datasets (one science exam and two English reading comprehension exams). They designed a reference prompt which asked GPT-3.5-turbo-0613 to role-play a student with one of the five proficiency levels and then generate an item response, an explanation and an estimated item difficulty. The experiments were conducted with zero-shot and temperature of 0 to reduce variability in LLMs' outputs. The results showed that GPT-3.5-turbo-0613 can simulate students at different proficiency levels to some extent, as the student level increased, the MC item response accuracy generally increased, and this pattern was observed on all three datasets. However, the reference prompt did not generalize well to GPT-3.5-turbo-1106 and GPT-4-1104-preview. In general, each LLM needed its own prompt engineering. Further, they found that the difficulty level predicted by the LLMs were not accurate enough, with Mean Absolute Percentage Error (MAPE) ranging from 12.81 to 32.70. Park et al. (2024) proposed the LLMs are Students at Various Levels (LLaSA) framework and used 65 LLMs from the HuggingFace Transformers library and the OpenAI API, including Llama, GPT-3.5, Mistral, to represent a range of student proficiency levels. Using item responses from LLMs, they used the Rasch model to estimate the item difficulty. In the zero-shot setting, this approach achieved a correlation of 0.35 with item difficulty estimates from IRT calibration. Zotos et al. (2025) combined textual representations of items, including TF-IDF and BERT embeddings, with LLM uncertainty features, including first-token probability and choice-order sensitivity to train a Random Forest regressor to predict Cambridge Multiple-Choice Questions Reading test items' p-values. The uncertainty features were extracted from nine open-source LLMs including models from Phi-3.5, Llama, Qwen2.5, and Yi-34B-Chat.

Their results showed that incorporating LLM uncertainty improved prediction accuracy over text-only baselines, achieving the lowest RMSE of 0.1651. Li et al. (2025) fine-tuned small language models and LLMs for difficulty prediction for the USMLE MC items from BEA 2024 Shared Task. They experimented with different LLM-related strategies, including the use of GPT-4-generated rationales as augmented data for fine-tuning encoder-only language models, as well as direct difficulty prediction and in-context learning with GPT-4. An ensemble model of two BERT- models with different data augmentation strategies yielded the lowest RMSE of 0.2926. However, LLMs didn't outperform the small language models. GPT-4 struggled with item difficulty prediction likely due to limited training data and the absence of explicit difficulty-related context.

## 3. Methods

### 3.1 Dataset

To investigate LLMs' capabilities in predicting item difficulty levels, this study used 1,443 MC items from a large-scale reading and writing test, designed to assess students' literacy and verbal reasoning skills in 4 content domains: Craft and Structure, Information and Ideas, Standard English Conventions, and Expression of Ideas for college admission. Both structural metadata and textual content are available for each item. Item metadata includes domain and skill labels and difficulty levels: easy, medium, and hard. Item text contains a short passage, an item stem, four answer options, the correct answer, and a rationale explaining why an option is correct and the other options are incorrect. Table 1 summarizes cross-tabulation of item distributions in terms of domain, skill, and item difficulty levels. In general, a relatively even distribution of each skill across different difficulty levels was observed, with no class imbalance issue.

**Table1.**

*Cross-tabulation of Items by domains, skills, and difficulty levels*

| Domain | Skill | Easy | Medium | Hard | Total |
|---|---|---|---|---|---|
| Craft and Structure | Words in Context | 116 | 46 | 43 | 205 |
| | Text Structure and Purpose | 36 | 49 | 32 | 117 |
| | Cross-Text Connections | 15 | 19 | 19 | 53 |
| Information and Ideas | Central Ideas and Details | 31 | 42 | 33 | 106 |
| | Inferences | 19 | 36 | 51 | 106 |
| | Command of Evidence | 61 | 71 | 88 | 220 |
| Standard English Conventions | Boundaries | 52 | 47 | 68 | 167 |
| | Form, Structure, and Sense | 84 | 38 | 38 | 160 |
| Expression of Ideas | Rhetorical Synthesis | 35 | 90 | 36 | 161 |
| | Transitions | 66 | 53 | 29 | 148 |
| Total | | 515 | 491 | 437 | 1443 |

To compare models, the dataset was split into training, validation, and test sets using a 70/15/15 ratio, resulting in 1,010, 216, and 217 items, respectively.

3.2 Models

To evaluate LLMs' performance in predicting item difficulty levels, multiple LLMs including GPT-3.5-turbo, GPT-4o, GPT-4o-mini, GPT-4.1, GPT-4.1-nano, GPT-4.1-mini, GPT-5, GPT-5.2, GPT-5.4, DeepSeek, Llama, Phi, and Qwen, were prompted with different strategies. Their performance was compared with supervised machine learning, and encoder-only language models.

### 3.2.1 LLMs

Two main prompting strategies were explored in this study: direct prediction with zero-shot and few-shot prompting, and incorporated test-taker proficiency prediction with persona-based prompts. For direct prediction, this study experimented with multiple GPT models including GPT-3.5-turbo, GPT-4o, GPT-4o-mini, GPT-4.1, GPT-4.1-nano, GPT-4.1-mini, GPT-5, GPT-5.2, GPT-5.4 using both zero-shot and few-shots prompting for item difficulty prediction. This comparison aimed to examine whether in-context learning improved LLMs' prediction performance. For zero-shot prompting, LLMs were asked to predict item difficulty based only on task instructions and item content. An example prompt is presented in Appendix A. For few-shot prompting, a small number of example items with difficulty labels from the training/validation set were provided to LLMs with additional guidance. As illustrated in Figure 1, three different shots-selection strategies were experimented including Fixed-3-shots, Fixed-9-shots, and Adaptive-3-shots.

- Fixed-3-shots used three examples within the same skill, consisting of one example from each difficulty level.
- Fixed-9-shots similarly used nine examples, but with 3 examples from each difficulty level.
- Adaptive-3-shots selected examples dynamically for each item within the same skill, the nearest neighbor of the item was identified using embedding-based similarity from the merged training and validation sets, with embeddings extracted using the Multilingual-E5-large-instruct model (L. Wang et al., 2024), and finally 3 examples were selected, one from each difficulty level.

**Figure 1.**

*Few-shot Examples Selection Strategies*

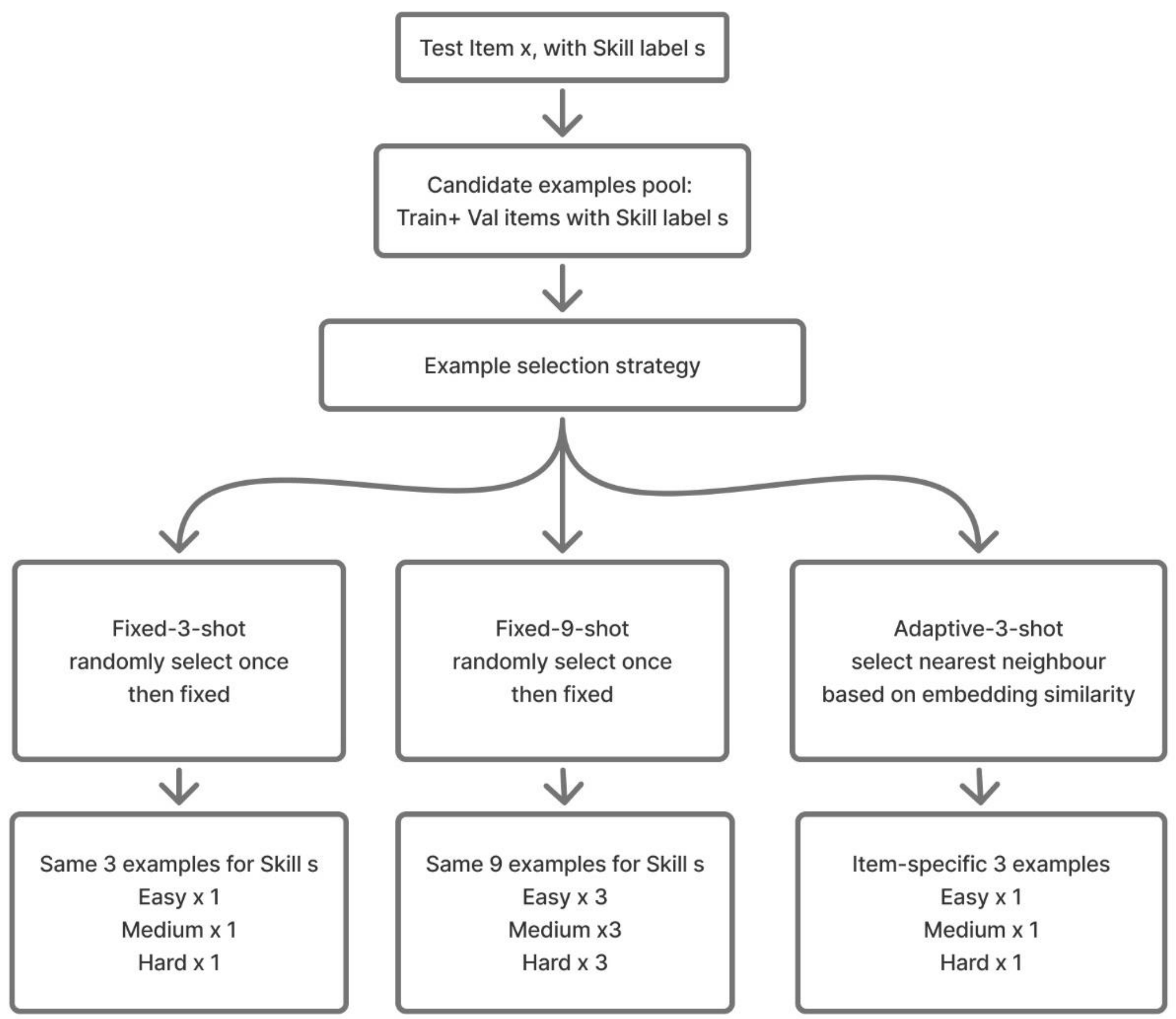

The effect of LLMs parameter settings was also examined. For non-reasoning models, including GPT-3.5-turbo, GPT-4, GPT-4o, and GPT-4.1, temperature was set to 0 representing a highly constrained setting to reduce randomness, as opposed to the default setting of 1. For reasoning models, such as GPT-5, reasoning effort was varied to adjust the extent of reasoning before response generation. Four reasoning effort settings: minimal, low, medium and high were tested for GPT-5, where medium is the default reasoning effort setting. For the subsequent comparison with GPT-5.2 and GPT-5.4, reasoning effort was fixed at medium for all models to ensure a fair comparison.

In addition, persona was used as another prompting strategy to investigate LLMs from multiple families, including GPT-3.5-turbo, GPT-4.1, GPT-4.1-mini, GPT-4o, GPT-4o-mini, GPT-5, GPT-o4-mini, Llama 2, Llama 3.1, Qwen 2.5, Qwen 3, Phi 3, Phi 3.5, Phi 4, DeepSeek-R1, QWQ-32B, and R1-Qwen32. Under this condition, the model was prompted to adopt a specific role to mimicking different ability levels in predicting item difficulty. This analysis used 1,338 items without figures. A no-persona condition was included as the baseline, with three persona conditions in which the model was instructed to respond as a weak, average, or strong student (full prompts are provided in Appendix B).

3.2.2 Supervised machine learning models

The supervised machine learning methods used features extracted from each item's textual content. Linguistic features were extracted using the Python package LFTK (Linguistic Feature ToolKit; Lee & Lee, 2023). The LFTK package contains 220 engineered features. Features were extracted separately for item information and rationale, resulting in a total of 440 linguistic features per item.

Optuna (Akiba et al., 2019) was then used as an automatic hyperparameter optimization framework to jointly search the feature selection strategy (no selection, ANOVA F-value ranking, mutual information ranking, L1-regularized logistic regression, and random-forest importance scoring), the percentage of selected features (5%, 10%, 20%, 40%, 60%, 80%, and 100% of 440 features), and classifier hyperparameter.

The experimented classifiers included linear SVM, Radial Basis Function (RBF) SVM,

logistic regression, random forest, decision tree, k-Nearest Neighbors (KNN), Light Gradient Boosting Machine (LightGBM), and Multinomial Naive Bayes. Each trial was evaluated by 5-fold cross-validation on the training/validation set, and Quadratic Weighted Kappa (QWK) was used as the main selection metric. After optimization, the best configuration was used to train the model and the trained model is tested on the test set.

### 3.2.3 Encoder-only language models

Using the same experimental setup described above, several encoder-only language models, including BERT, ConvBERT, DeBERTa, ELECTRA, and RoBERTa variants, were trained and evaluated for item difficulty prediction. BERT (Devlin et al., 2019) is a foundational model in modern NLP research, introducing a deep bidirectional transformer architecture that learn bidirectional contextual representations. RoBERTa (Liu et al., 2019) keeps the same general architecture as BERT but improves performance through stronger training strategies, including more data, longer training, dynamic masking, and removing the next sentence prediction objective. ConvBERT (Jiang et al., 2020) incorporates convolutional operations into the Transformer structure to better capture local contextual patterns while improving computational efficiency. DeBERTa (He et al., 2021) further improves BERT by introducing a disentangled attention mechanism that separates content and position representations, allowing the model to capture word relationships more precisely. ELECTRA (Clark et al., 2020) changes the pre-training objective and introduces the replaced token detection, making learning more sample efficient. This study experimented with BERT-base, ConvBERT-base, DeBARTa-v3-large, ELECTRA-base, RoBERTa-large. Both linear and cosine annealing schedulers were explored for model training

3.2.4 Evaluation of Model Performances

All model performances were evaluated using four metrics: QWK, accuracy, class-specific F1 scores, and Spearman rank-order correlation for three levels of item difficulty classification. Among these metrics, the QWK coefficient was adopted as the primary evaluation metric because it extends Cohen's Kappa and applies weighted penalties to disagreements, penalizing larger disagreements more heavily than smaller ones. The three difficulty levels, Easy, Medium, and Hard, were treated as ordinal labels and encoded as 0, 1, and 2, respectively. Let $O_{ij}$ denote the observed confusion matrix cell value, where $O_{ij}$ is the number of samples with ground-truth label $i$ and predicted label $j$. Let $E_{ij}$ denote the expected confusion matrix cell value under independent ground-truth and predicted label distributions. The quadratic weight between classes $i$ and $j$ was defined as

$$w_{ij} = \frac{(i-j)^2}{(K-1)^2} \tag{1}$$

where $K$=3 is the number of difficulty levels. QWK was calculated as

$$\kappa = 1 - \frac{\sum_{i=0}^{K-1}\sum_{j=o}^{K-1} w_{ij}O_{ij}}{\sum_{i=0}^{K-1}\sum_{j=0}^{K-1} w_{ij}E_{ij}} \tag{2}$$

The expected matrix $E_{ij}$ was computed as

$$E_{ij} = \frac{n_i m_j}{N} \tag{3}$$

where

$$n_i = \sum_{j=0}^{K-1} O_{ij} \tag{4}$$

and

$$m_j = \sum_{i=0}^{K-1} O_{ij} \tag{5}$$

are the marginal counts of the ground-truth and predicted labels, respectively. Accuracy was used as a straightforward measure of overall classification performance, as shown in Equation 6.

$$Accuracy = \frac{TP_{Easy} + TP_{Medium} + TP_{Hard}}{N} \quad (6)$$

where $TP_{Easy}$, $TP_{Medium}$, and $TP_{Hard}$ denote the numbers of correctly predicted samples for the Easy, Medium and Hard levels, respectively, and $N$ denotes the total number of samples. Class-specific F1 scores were computed for each difficulty level to measure category-level performance (Sokolova & Lapalme, 2009), as shown in Equation 7.

$$F1_c = \frac{2TP_c}{2TP_c + FP_c + FN_c} \quad (7)$$

where $TP_c$, $FP_c$, and $FN_c$ denote the number of true positives, false positives and false negatived for difficulty level $c$, respectively. Spearman rank-order correlation was computed to capture the ordinal association between predicted and actual difficulty levels (Li, Chen, et al., 2025), as shown in Equation 8.

$$\rho = \frac{\sum_{i=1}^{n}(R_i - \bar{R})(S_i - \bar{S})}{\sqrt{\sum_{i=1}^{n}(R_i - \bar{R})^2 \sum_{i=1}^{n}(S_i - \bar{S})^2}} \quad (8)$$

3.2.5 PCA Visualization of Item Embeddings

To better understand item text representation, embeddings were generated using the Multilingual-E5-large-instruct (L. Wang et al., 2024) for each item. Three sets of embeddings were compiled respectively for item content, the rationale and combination of item information and the rationale for all items. Principal component analysis (PCA) was then applied for dimension reduction for each embedding set. PCA results help to understand the challenges or easiness in item difficulty level prediction per se.

## 4. Results

### 4.1 LLM-based prediction

Table 2 presents the model performance comparison under zero-shot prompting, with reasoning effort fixed at medium (defaulting setting for GPT-5) for reasoning models and temperature fixed at 1 (default setting) for non-reasoning models. Among these models, GPT-4.1 achieved the highest QWK of 0.447 and the highest accuracy of 0.535, outperforming GPT-5.4, the latest reasoning model, and performed particularly well on predicting items of medium difficulty. The reasoning models (GPT-5, GPT-5.2 and GPT-5.4) performed well on easy items, achieving F1 scores above 0.57. However, their performance dropped sharply on medium difficulty level prediction and became extremely poor on hard difficulty level prediction, where the highest F1 score was only 0.088.

**Table 2.**

*Performance of GPT Models under Zero-Shot Prompting*

| LLM | QWK | Accuracy | Spearman Correlation | F1_Easy | F1_Medium | F1_Hard |
|---|---|---|---|---|---|---|
| GPT 3.5-turbo | 0.328 | 0.475 | 0.343 | 0.558 | 0.487 | 0.319 |
| GPT 4o | 0.395 | 0.461 | 0.453 | 0.298 | 0.491 | **0.533** |
| GPT 4o-mini | 0.26 | 0.429 | 0.326 | 0.163 | 0.522 | 0.424 |
| GPT 4.1 | **0.447** | **0.535** | **0.503** | 0.612 | **0.578** | 0.296 |
| GPT 4.1-nano | 0.123 | 0.369 | 0.229 | 0.026 | 0.509 | 0.208 |
| GPT 4.1-mini | 0.395 | 0.465 | 0.441 | 0.587 | 0.466 | 0.231 |
| GPT 5 | 0.198 | 0.392 | 0.457 | 0.597 | 0.144 | 0.000 |
| GPT 5.2 | 0.254 | 0.419 | 0.481 | **0.613** | 0.203 | 0.088 |
| GPT 5.4 | 0.197 | 0.373 | 0.404 | 0.576 | 0.075 | 0.086 |

*Note*. The bold value indicates the highest score for each metric.

The performance of the GPT models under different parameter and prompting combinations is summarized in Table 3. GPT-4.1 with temperature of 0 and zero-shot prompting achieved the best overall performance, with a QWK of 0.578 and an accuracy of 0.618. The reasoning model GPT-5 benefited from the examples in the prompts. Compared with zero-shot prompting, both fixed-shot and adaptive-shot prompting produced higher QWK and accuracy. Furthermore, providing more examples appeared to improve performance, with fixed-9-shot condition outperforming the fixed-3-shot. By contrast, the non-reasoning model, GPT-4.1 achieved its best performance under zero-shot prompting, and the fixed-9-shot condition performed worse than fixed-3-shot. The reasoning effort for GPT-5 was compared in the zero-shot prompting. Among the four available settings, minimal effort achieved the best performance with QWK of 0.41. For non-reasoning models, including GPT-3.5-turbo, GPT4, GPT-4o, and GPT-4.1, the effect of temperature was not consistent across models and the prompting condition. For GPT-3.5-turbo with a temperature of 0 and fixed-9-shots, all items were classified as medium, hence, Spearman’s rho was undefined. In general, temperature 0 performed better under zero-shot prompting, whereas temperature 1 under few-shot prompting performed better for GPT-3.5-turbo, GPT-4, and GPT-4.1, but not for GPT-4o.

**Table 3.**

*Performance of GPT Models under Different Parameter and Prompting Combinations*

| Prompting Strategy | LLM | Parameter | QWK | Accuracy | Spearman Correlation | F1_Easy | F1_Medium | F1_Hard |
|---|---|---|---|---|---|---|---|---|
| Zero-shot | GPT-4.1 | 0 | **0.578** | **0.618** | **0.600** | 0.661 | **0.618** | **0.560** |
| | GPT-4.1 | 1 | 0.447 | 0.535 | 0.503 | 0.612 | 0.578 | 0.296 |
| | GPT-5 | minimal | 0.410 | 0.498 | 0.518 | 0.651 | 0.387 | 0.308 |
| | GPT-5 | low | 0.294 | 0.429 | 0.462 | 0.622 | 0.303 | 0.060 |
| | GPT-5 | medium | 0.198 | 0.392 | 0.457 | 0.597 | 0.144 | 0.000 |
| | GPT-5 | high | 0.217 | 0.396 | 0.472 | 0.597 | 0.128 | 0.060 |
| Fixed-3-shots | GPT-4.1 | 0 | 0.512 | 0.544 | 0.584 | **0.698** | 0.429 | 0.418 |
| | GPT-5 | medium | 0.221 | 0.415 | 0.410 | 0.599 | 0.220 | 0.087 |
| Adaptive-3-shots | GPT-4.1 | 0 | 0.479 | 0.502 | 0.584 | 0.680 | 0.380 | 0.317 |
| | GPT-5 | medium | 0.251 | 0.401 | 0.443 | 0.586 | 0.140 | 0.169 |
| Fixed-9-shots | GPT-3.5-turbo | 0 | 0.000 | 0.346 | — | 0.000 | 0.514 | 0.000 |
| | GPT-3.5-turbo | 1 | 0.018 | 0.355 | 0.045 | 0.000 | 0.518 | 0.085 |
| | GPT-4 | 0 | 0.285 | 0.452 | 0.362 | 0.487 | 0.540 | 0.085 |
| | GPT-4 | 1 | 0.313 | 0.475 | 0.379 | 0.487 | 0.557 | 0.187 |
| | GPT-4.1 | 0 | 0.468 | 0.53 | 0.556 | 0.688 | 0.451 | 0.321 |
| | GPT-4.1 | 1 | 0.474 | 0.521 | 0.571 | 0.695 | 0.429 | 0.296 |
| | GPT-4o | 0 | 0.367 | 0.447 | 0.440 | 0.592 | 0.460 | 0.087 |
| | GPT-4o | 1 | 0.329 | 0.498 | 0.389 | 0.620 | 0.527 | 0.160 |

| GPT-5 | medium | 0.305 | 0.438 | 0.496 | 0.625 | 0.240 | 0.167 |
|---|---|---|---|---|---|---|---|

*Note*. Parameter denotes temperature for non-reasoning models (including GPT-3.5-turbo, GPT4, GPT-4o, and GPT-4.1) and reasoning effort for reasoning models (GPT-5). "—" denotes an undefined Spearman Correlation. The bold value indicates the highest score for each metric.

To better understand how models differ in their prediction performance, we selected four models based on the performance under zero-shot condition and analyzed their confusion matrices (see Table 4). These models represented the best-performing, an intermediate-performing, and the lowest-performing conditions, illustrating prediction patterns across the Easy, Medium, and Hard. GPT-4.1 showed the most balanced performance among these models, and it more accurately predicted items with medium difficulty, with 85.33% of actual medium difficult items predicted as Medium. GPT-3.5-turbo performed moderately. GPT-5 and GPT-5.4 showed a strong tendency toward predicting difficult items as easy or medium difficult items, suggesting a bias to underestimate item difficulty given their higher capabilities compared with the other two lower general capability models. Though all models struggled with identifying hard items, more competent models such as GPT-5.2 and GPT-5.4 demonstrated they could not perceive the challenges students may face with difficult items.

**Table 4.**

*Confusion Matrices for Difficulty Prediction across GPT models under Zero-Shot Prompting*

| LLMs | | Predicted Difficulty | | |
|---|---|---|---|---|
| | Known Difficulty | Easy | Medium | Hard |
| GPT-4.1 | Easy | **54.55%** | 44.16% | 1.30% |
| | Medium | 6.67% | **85.33%** | 8.00% |
| | Hard | 4.62% | 52.31% | **43.08%** |
| GPT-3.5-turbo | Easy | **53.25%** | 44.16% | 2.60% |
| | Medium | 21.33% | **62.47%** | 16.00% |
| | Hard | 20.00% | 56.92% | **23.08%** |
| GPT-5 | Easy | **100%** | 0 | 0 |
| | Medium | 89.33% | **10.67%** | 0 |
| | Hard | 56.92% | 43.08% | **0** |
| GPT-5.4 | Easy | **96.10%** | 3.90% | 0 |
| | Medium | 92.00% | **5.33%** | 2.67% |
| | Hard | 56.92% | 38.46% | **4.62%** |

The results of the experiments with different persona promptings are summarized in Table 5 and 6 for the overall and level-specific prediction accuracy for different LLMs under the baseline, weak, average and good-level student persona settings. Overall, the effects of persona prompting are mixed and generally limited on most metrics. GPT-4.1 with good-level student persona prompting achieved the best accuracy and Spearman correlation, while GPT-5 under weak-level student persona prompting yielded the highest QWK score. One notable pattern is that the weak-level student persona prompting tended to improve F1-Hard score and reduce F1-Easy score for most LLMs, especially for the GPT family models. For example, GPT-4.1 achieved a substantial improvement on F1-Hard score, increasing from 0.189 under the baseline setting to 0.575 under the weak persona setting. Smaller models, such as Liama2-7B and Liama2-13B, appeared less responsive to persona changes.

**Table 5.**

*Performance of LLMs under Different Persona Conditions*

| LLM | Baseline | | | Weak | | | Average | | | Good | | |
|---|---|---|---|---|---|---|---|---|---|---|---|---|
| | QWK | Accuracy | Spearman Correlation | QWK | Accuracy | Spearman Correlation | QWK | Accuracy | Spearman Correlation | QWK | Accuracy | Spearman Correlation |
| GPT-3.5-Turbo | 0.204 | 0.432 | 0.259 | 0.220 | 0.423 | 0.228 | 0.168 | 0.396 | 0.194 | 0.143 | 0.391 | 0.155 |
| GPT-4.1 | 0.339 | 0.460 | **0.492** | 0.389 | 0.452 | 0.469 | **0.400** | **0.474** | 0.465 | 0.404 | **0.497** | **0.512** |
| GPT-4.1-mini | **0.415** | 0.491 | 0.491 | 0.404 | 0.478 | 0.424 | 0.213 | 0.415 | 0.304 | 0.386 | 0.475 | 0.436 |
| GPT-4o | 0.315 | 0.456 | 0.381 | 0.298 | 0.432 | 0.343 | 0.237 | 0.415 | 0.312 | 0.359 | 0.472 | 0.416 |
| GPT-4o-mini | 0.213 | 0.401 | 0.264 | 0.237 | 0.401 | 0.274 | 0.130 | 0.372 | 0.194 | 0.211 | 0.392 | 0.248 |
| GPT-5 | 0.136 | 0.377 | 0.379 | **0.433** | 0.490 | **0.496** | 0.292 | 0.427 | **0.499** | 0.124 | 0.373 | 0.353 |
| GPT-o4-mini | 0.339 | 0.459 | 0.461 | 0.376 | 0.467 | 0.409 | 0.367 | 0.458 | 0.449 | 0.322 | 0.445 | 0.423 |
| DeepSeek-R1 | 0.400 | **0.494** | 0.502 | 0.404 | **0.490** | 0.448 | 0.286 | 0.445 | 0.384 | **0.420** | 0.496 | 0.494 |
| Llama2-13B | 0.112 | 0.357 | 0.125 | 0.078 | 0.359 | 0.087 | 0.059 | 0.348 | 0.070 | 0.083 | 0.358 | 0.092 |
| Llama2-7B | 0.063 | 0.346 | 0.090 | 0.070 | 0.350 | 0.083 | 0.040 | 0.340 | 0.054 | 0.044 | 0.328 | 0.054 |
| Llama3.1-8B | 0.216 | 0.397 | 0.233 | 0.229 | 0.394 | 0.248 | 0.171 | 0.381 | 0.207 | 0.216 | 0.413 | 0.242 |
| Phi3 | 0.090 | 0.387 | 0.091 | 0.105 | 0.369 | 0.111 | 0.043 | 0.359 | 0.049 | 0.142 | 0.391 | 0.142 |
| Phi3.5 | 0.114 | 0.374 | 0.127 | 0.120 | 0.382 | 0.127 | 0.137 | 0.380 | 0.146 | 0.110 | 0.379 | 0.112 |
| Phi4 | 0.344 | 0.451 | 0.382 | 0.297 | 0.437 | 0.345 | 0.319 | 0.457 | 0.386 | 0.339 | 0.472 | 0.407 |
| Qwen2.5-32B | 0.247 | 0.431 | 0.301 | 0.278 | 0.440 | 0.305 | 0.240 | 0.431 | 0.303 | 0.288 | 0.455 | 0.340 |
| Qwen2.5-7B | 0.169 | 0.419 | 0.216 | 0.204 | 0.430 | 0.278 | 0.192 | 0.423 | 0.247 | 0.203 | 0.452 | 0.252 |
| Qwen3-32B-NR | 0.176 | 0.383 | 0.236 | 0.266 | 0.409 | 0.316 | 0.139 | 0.380 | 0.222 | 0.249 | 0.410 | 0.300 |
| Qwen3-32B-R | 0.259 | 0.436 | 0.328 | 0.223 | 0.404 | 0.296 | 0.077 | 0.361 | 0.157 | 0.255 | 0.441 | 0.321 |
| Qwen3-8B | 0.149 | 0.395 | 0.200 | 0.198 | 0.405 | 0.238 | 0.183 | 0.394 | 0.236 | 0.178 | 0.411 | 0.218 |
| QWQ-32B | 0.351 | 0.469 | 0.431 | 0.288 | 0.432 | 0.361 | 0.264 | 0.435 | 0.353 | 0.351 | 0.471 | 0.438 |
| R1-Qwen32B | 0.192 | 0.418 | 0.249 | 0.191 | 0.411 | 0.237 | 0.132 | 0.393 | 0.185 | 0.223 | 0.421 | 0.279 |

*Note*. Default parameter settings were used for all LLMs. Qwen3-32B-R refers to the reasoning version, and Qwen3-32B-NR refers to the non-reasoning version. The bold values indicate the highest scores for each metric.

**Table 6.**

*F1 scores of LLMs under Different Persona Conditions*

| LLM | Baseline | | | Weak | | | Average | | | Good | | |
|---|---|---|---|---|---|---|---|---|---|---|---|---|
| | F1_Easy | F1_Medium | F1_Hard | F1_Easy | F1_Medium | F1_Hard | F1_Easy | F1_Medium | F1_Hard | F1_Easy | F1_Medium | F1_Hard |
| GPT-3.5-Turbo | 0.559 | 0.423 | 0.137 | 0.447 | 0.462 | 0.309 | 0.383 | 0.488 | 0.132 | 0.355 | 0.477 | 0.223 |
| GPT-4.1 | 0.641 | 0.314 | 0.189 | 0.441 | 0.290 | **0.575** | 0.617 | 0.471 | 0.173 | **0.663** | 0.390 | 0.274 |
| GPT-4.1-mini | 0.637 | 0.509 | 0.130 | 0.426 | 0.482 | 0.522 | 0.383 | 0.525 | 0.025 | 0.583 | 0.497 | 0.210 |
| GPT-4o | 0.566 | 0.506 | 0.062 | 0.433 | 0.501 | 0.215 | 0.391 | 0.514 | 0.084 | 0.570 | 0.513 | 0.158 |
| GPT-4o-mini | 0.285 | 0.504 | 0.222 | 0.215 | 0.470 | 0.418 | 0.168 | 0.503 | 0.142 | 0.214 | 0.482 | **0.345** |
| GPT-5 | 0.573 | 0.088 | 0.000 | **0.642** | 0.410 | 0.318 | **0.635** | 0.275 | 0.015 | 0.567 | 0.083 | 0.000 |
| GPT-o4-mini | 0.644 | 0.426 | 0.015 | 0.511 | 0.505 | 0.304 | 0.602 | 0.491 | 0.039 | 0.617 | 0.428 | 0.020 |
| DeepSeek-R1 | **0.672** | 0.486 | 0.072 | 0.468 | **0.539** | 0.389 | 0.467 | **0.537** | 0.058 | 0.650 | 0.504 | 0.156 |
| Llama2-13B | 0.232 | 0.461 | 0.233 | 0.266 | 0.466 | 0.202 | 0.170 | 0.474 | 0.190 | 0.242 | 0.459 | 0.263 |
| Llama2-7B | 0.120 | 0.486 | 0.124 | 0.237 | 0.469 | 0.148 | 0.119 | 0.483 | 0.123 | 0.160 | 0.454 | 0.167 |
| Llama3.1-8B | 0.309 | 0.456 | **0.367** | 0.348 | 0.461 | 0.293 | 0.211 | 0.488 | **0.257** | 0.354 | 0.498 | 0.261 |
| Phi3 | 0.446 | 0.394 | 0.282 | 0.454 | 0.342 | 0.275 | 0.364 | 0.439 | 0.171 | 0.441 | 0.382 | 0.330 |
| Phi3.5 | 0.406 | 0.440 | 0.150 | 0.446 | 0.403 | 0.232 | 0.366 | 0.457 | 0.210 | 0.354 | 0.441 | 0.294 |
| Phi4 | 0.546 | 0.469 | 0.224 | 0.580 | 0.428 | 0.172 | 0.600 | 0.477 | 0.082 | 0.617 | 0.448 | 0.191 |
| Qwen2.5-32B | 0.477 | 0.511 | 0.071 | 0.431 | 0.505 | 0.278 | 0.477 | 0.516 | 0.025 | 0.543 | 0.511 | 0.106 |
| Qwen2.5-7B | 0.445 | 0.519 | 0.000 | 0.569 | 0.434 | 0.020 | 0.449 | 0.520 | 0.000 | 0.530 | **0.525** | 0.034 |
| Qwen3-32B-NR | 0.259 | 0.498 | 0.142 | 0.191 | 0.483 | 0.425 | 0.127 | 0.514 | 0.182 | 0.234 | 0.503 | 0.339 |
| Qwen3-32B-R | 0.434 | **0.529** | 0.107 | 0.232 | 0.512 | 0.250 | 0.098 | 0.509 | 0.072 | 0.410 | 0.536 | 0.150 |
| Qwen3-8B | 0.362 | 0.508 | 0.010 | 0.417 | 0.494 | 0.083 | 0.373 | 0.498 | 0.047 | 0.405 | 0.509 | 0.075 |
| QWQ-32B | 0.599 | 0.515 | 0.025 | 0.469 | 0.512 | 0.077 | 0.478 | 0.527 | 0.000 | 0.624 | 0.499 | 0.020 |
| R1-Qwen32B | 0.408 | 0.526 | 0.013 | 0.373 | 0.518 | 0.067 | 0.278 | 0.523 | 0.051 | 0.455 | 0.514 | 0.032 |

*Note*. Default parameter settings were used for all LLMs. Qwen3-32B-R refers to the reasoning version, and Qwen3-32B-NR refers to the non-reasoning version. The bold values indicate the highest scores for each metric.

4.2 Supervised machine learning

The performance of the feature-based supervised machine learning models is summarized in Table 7. Overall, linear SVM and RBF SVM had the top performances across the metrics, and linear SVM obtained the highest QWK of 0.549 with 40% linguistic features. For different difficulty levels, all models performed better on the Easy and Hard categories than on the Medium. For example, linear SVM achieved an F1 score of 0.671 for Easy and 0.635 for Hard, but only 0.478 for Medium. This pattern may suggest that medium difficulty items were the most challenging to classify due to the share features with both easier and harder items.

**Table 7.**

*Performance for Supervised Machine Learning*

| Model | % Selected Features | QWK | Accuracy | Spearman Correlation | F1_Easy | F1_Medium | F1_Hard |
|---|---|---|---|---|---|---|---|
| Linear SVM | 40% | **0.549** | 0.599 | 0.548 | 0.671 | 0.478 | **0.635** |
| RBF SVM | 80% | 0.547 | **0.608** | **0.558** | **0.709** | 0.507 | 0.581 |
| LightGBM | 40% | 0.531 | 0.590 | 0.532 | 0.658 | 0.519 | 0.593 |
| Logistic regression | 80% | 0.527 | 0.571 | 0.532 | 0.683 | 0.455 | 0.557 |
| Naïve Bayes | 10% | 0.494 | 0.535 | 0.537 | 0.657 | 0.368 | 0.520 |
| KNN | 40% | 0.454 | 0.581 | 0.466 | 0.667 | **0.521** | 0.523 |
| Random forest | 10% | 0.447 | 0.558 | 0.459 | 0.636 | 0.479 | 0.536 |
| Decision tree | 100% | 0.403 | 0.516 | 0.410 | 0.556 | 0.431 | 0.562 |

*Note*. The bold value indicates the highest score for each metric.

4.3 Encoder-only language models

The performance of the encoder-only language models is summarized in Table 8. Overall, ConvBERT achieved the best performance across the main evaluation metrics. ConvBERT with linear scheduler obtained the best QWK of 0.625 and the best Spearman correlation of 0.632. ConvBERT with cosine annealing achieved the highest accuracy of 0.631. Difficulty-level specific F1 scores varied across models. In general, models achieved highest F1 scores on easy items, while performance on medium items was relatively weak. And the results were consistently with machine learning results: models performed worse on medium items than on easy and hard items. The best F1 score for easy items was obtained by DeBERTa-v3-large with linear scheduler at 0.759, whereas ELECTRA-base with linear scheduler achieved the highest F1 score for medium items at 0.576. For hard items, DeBERTa-v3-large with cosine annealing performed the best, with an F1 score of 0.638. These results suggest that no single encoder-only language model dominated across all difficulty levels.

**Table 8.**

*Performance for Encoder-only Language Models*

| Model | QWK | Accuracy | Spearman Correlation | F1_Easy | F1_Medium | F1_Hard |
|---|---|---|---|---|---|---|
| BERT-linear | 0.476 | 0.566 | 0.481 | 0.629 | 0.432 | 0.613 |
| BERT-cosine | 0.546 | 0.562 | 0.553 | 0.710 | 0.355 | 0.567 |
| ConvBERT-linear | **0.625** | 0.599 | **0.632** | 0.714 | 0.503 | 0.584 |
| ConvBERT-cosine | 0.603 | **0.631** | 0.607 | 0.725 | 0.546 | 0.617 |
| DeBERTa-linear | 0.594 | 0.618 | 0.609 | **0.759** | 0.462 | 0.585 |
| DeBERTa-cosine | 0.599 | 0.599 | 0.608 | 0.672 | 0.503 | **0.638** |
| ELECTRA-linear | 0.524 | 0.590 | 0.558 | 0.701 | **0.576** | 0.427 |
| ELECTRA-cosine | 0.538 | 0.544 | 0.556 | 0.618 | 0.468 | 0.571 |
| RoBERTa-linear | 0.555 | 0.599 | 0.571 | 0.688 | 0.467 | 0.613 |
| RoBERTa-cosine | 0.533 | 0.604 | 0.536 | 0.687 | 0.514 | 0.597 |

*Note*. The bold value indicates the highest score for each metric.

4.4 PCA Visualization of Item Embeddings

Figures 2 to 4 present the PCA visualization of the item information embeddings, rationale embeddings, and combined item information and rationale embeddings, respectively. Across all three figures, the easy, medium and hard items were highly overlapping in the first two principal components. The rationale embeddings space showed a more visible global structure, with points forming two regions, however, the three difficulty levels items remained mixed.

**Figure 2.**

*PCA Visualization of Item Information Embeddings*

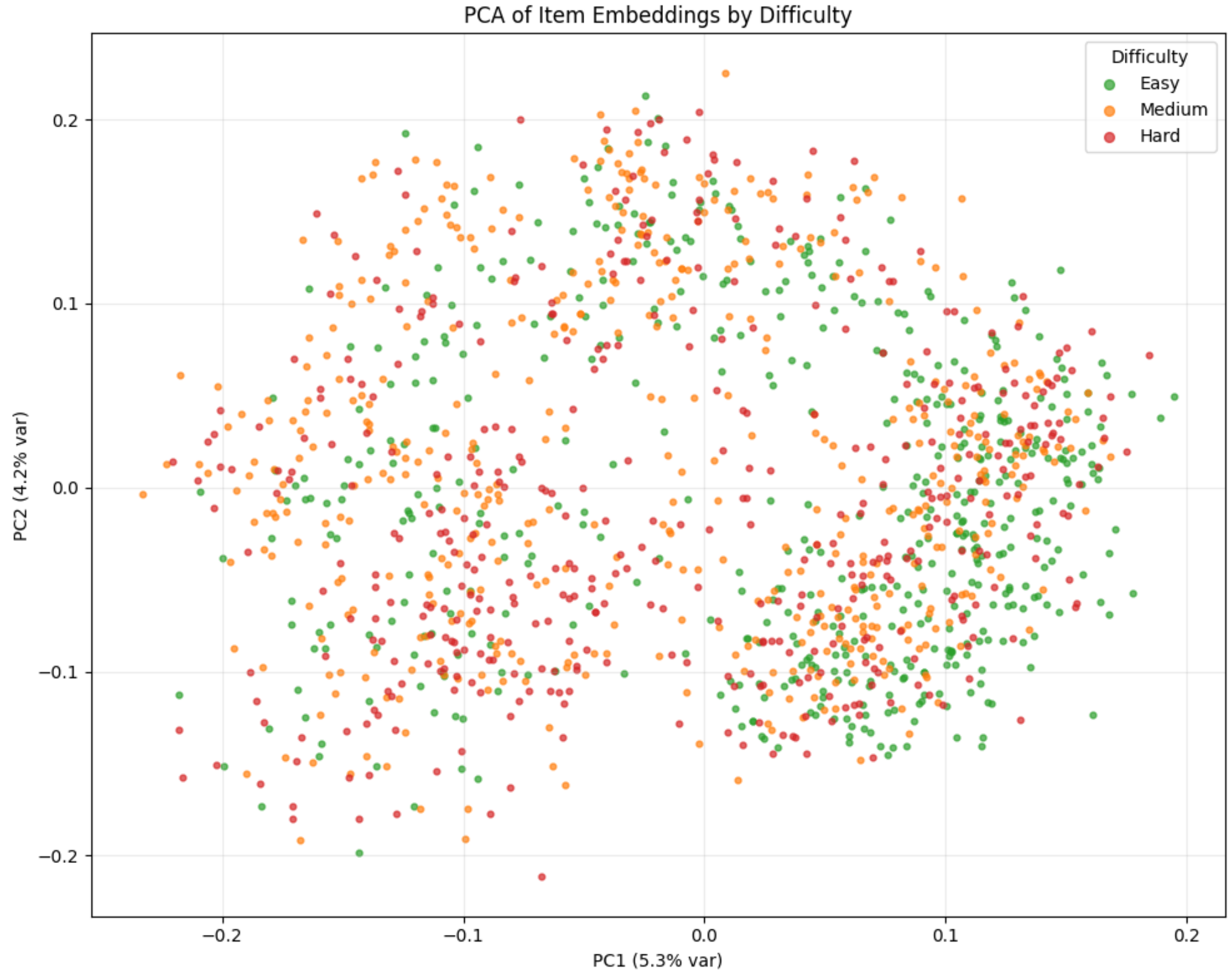

**Figure 3.**

*PCA Visualization of Rationale Embeddings*

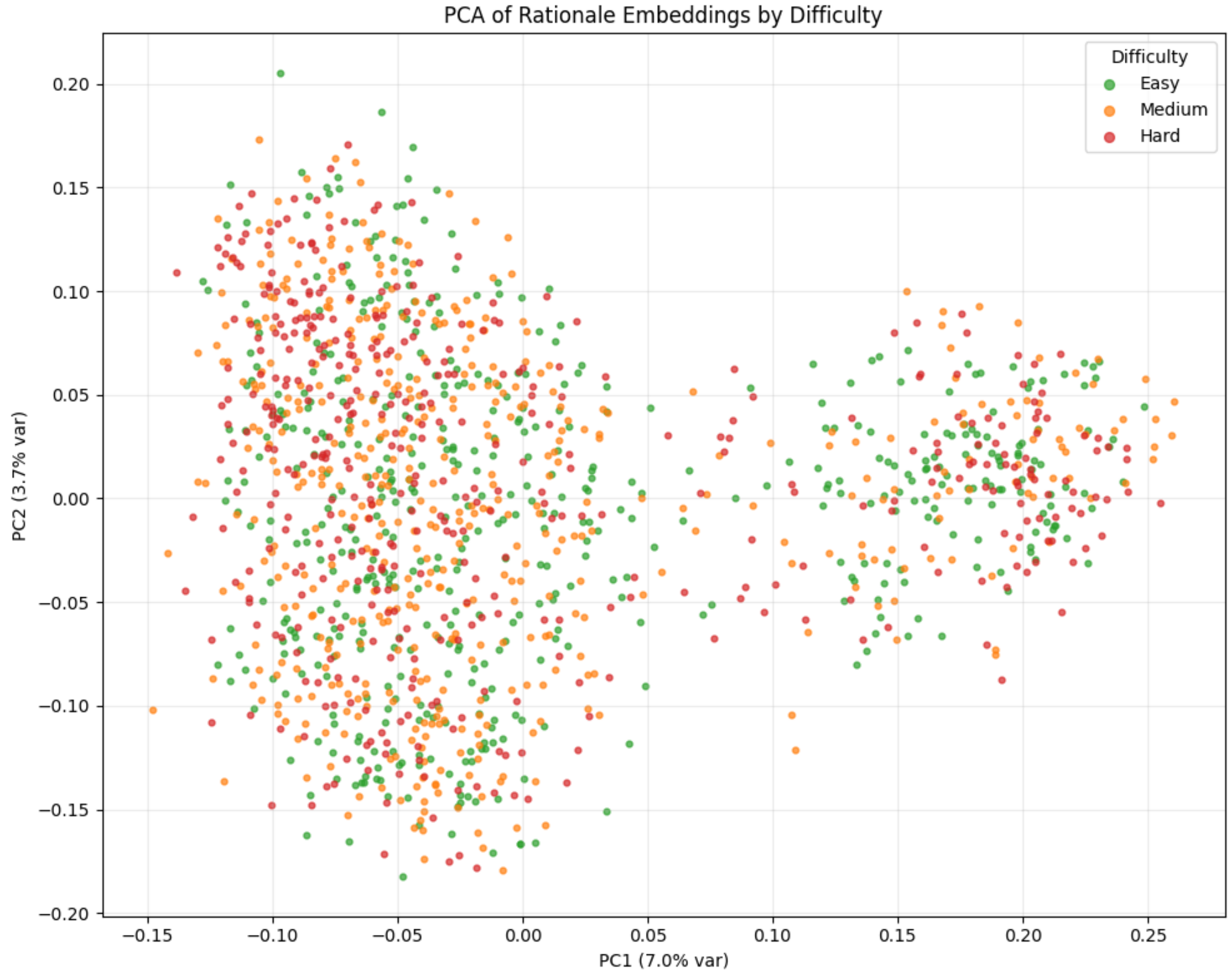

**Figure 4.**

*PCA Visualization of Combined Item Information and Rationale Embeddings*

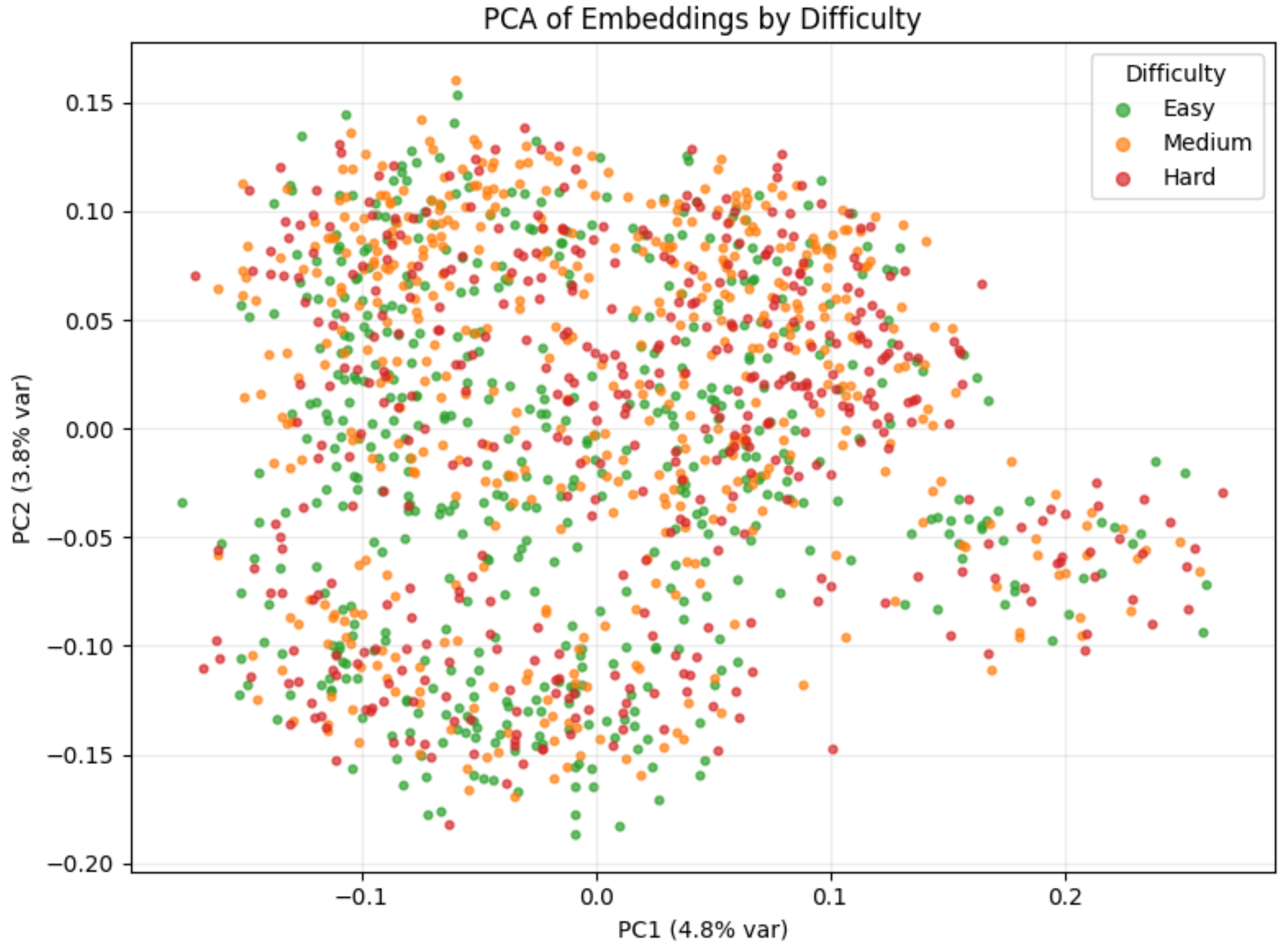

**5. Summary and Discussion**

This study explored different LLMs for item difficulty level prediction for items from a large-scale reading and writing test. Different prompting strategies and hyper-parameter settings were explored to better understand LLMs' capabilities in predicting item difficulty levels. LLMs' performance was compared with feature-based supervised machine learning models and encoder-only language models. The overall best performance was achieved by encoder-only language models, specifically the ConvBERT with linear scheduler, yielding a QWK of 0.625. Although the LLM-based approaches did not outperform this model, their performance varied across prompting strategies and model settings. Within the LLM-based approaches, Zero-Shot GPT 4.1 with temperature of 0 yielded the highest QWK of 0.578.

This comparative analysis of different difficulty prediction approaches revealed several key findings. First, the LLM-based predictions may be sensitive to prompt strategies and generation settings. Reasoning models, GPT5, with few-shot prompting tended to produce better performance, while non-reasoning models, GPT-3.5-turbo, GPT-4, GPT-4.1 and GPT-4o, with zero-shot prompting yielded better results. The effects of temperature were also mixed, with no consistent advantages for either deterministic (temperature=0) or more variable (temperature=1) generation settings across models. This finding aligns with Benedetto et al. (2024) in that the same prompt has limited generalizability even across GPT-series models, highlighting the need for model-specific prompt design and validation. Furthermore, all LLMs struggled at identifying hard items, especially with the latest more capable LLMs including GPT 5.4 and 5 showing a particularly strong tendency to underestimate item difficulty. Confusion matrix showed that these more capable models classified almost all hard items as easy or medium difficult items. This result is consistent with Li et al.(2025), that model-predicted difficulty distributions were compressed and skewed toward lower values, especially for advanced models, which may be due to a curse of knowledge.

Though Yi et al (2024) reported an accuracy of 0.806 for three level item difficulty level classification with only 263 items, our study only achieved an accuracy of 0.618 for GPT-4.1 with zero-shot and 0.631 for ConvBERT. One reason could be related to the content domain. Yi et al (2024) used college entrance exam math test items while our study used reading and writing items. Another possible explanation is that Yi et al (2024) crafted 8 features that may bear more construct relevant information related to item difficulty levels. This signifies that item semantic representation based on embeddings may not be sufficient to classify item difficulty levels accurately. The PCA visualization of item information embeddings further supported that the embeddings of items from different difficulty levels did not show enough separation for LLMs or encoder models to effectively cluster items. This also corroborates that item difficulty is not a mere textual property. Based on IRT, item difficulty (b) is a location index on the latent ability scale, reflecting the ability level at which examinees are expected to answer the item correctly with a certain probability (Baker, 2001). Item difficulty reflects the interaction relationship between item content and test-takers' latent ability (θ). For the same item, the difficulty levels could be different for different grades/ability groups like in vertical scaling. Text-based approaches to item difficulty prediction may miss the information from the ability facet.

How to effectively incorporate information from both facets: item characteristics and test-takers' ability, is a feasible direction for future research (Benedetto et al., 2024; Liu et al., 2025; Park et al., 2024). In this study, we made an initial attempt through persona prompting, where LLMs were instructed to respond as weak, average, or good students. However, this prompt strategy did not get a better prediction performance for GPT-based models and other LLMs, including Llama 2, Llama 3.1, Qwen 2.5, Qwen 3, Phi 3, Phi 3.5, Phi 4, DeepSeek-R1, QWQ-32B, and R1-Qwen32B. Weak student persona slightly mitigate the curse of knowledge, but the findings suggest only assigning simple ability level personas is insufficient for capturing the test-takers' ability for difficulty prediction. Other approaches should be more comprehensively explored to help LLMs get a clear understanding of item difficulty If LLMs cannot understand or predict item difficulty accurately, they may not be able to generate items with targeted item difficulty levels.

## References

Akiba, T., Sano, S., Yanase, T., Ohta, T., & Koyama, M. (2019). Optuna: A Next-generation Hyperparameter Optimization Framework. *Proceedings of the 25th ACM SIGKDD International Conference on Knowledge Discovery & Data Mining, KDD '19*, 2623–2631. https://doi.org/10.1145/3292500.3330701

AlKhuzaey, S., Grasso, F., Payne, T. R., & Tamma, V. (2024). Text-based Question Difficulty Prediction: A Systematic Review of Automatic Approaches. *International Journal of Artificial Intelligence in Education*, *34*(3), 862–914. https://doi.org/10.1007/s40593-023-00362-1

Baker, F. B. (2001). *The Basics of Item Response Theory. Second Edition*. For full text: http://ericae. https://eric.ed.gov/?id=ED458219

Benedetto, L. (2023). A Quantitative Study of NLP Approaches to Question Difficulty Estimation. In N. Wang, G. Rebolledo-Mendez, V. Dimitrova, N. Matsuda, & O. C. Santos (Eds.), *Artificial Intelligence in Education. Posters and Late Breaking Results, Workshops and Tutorials, Industry and Innovation Tracks, Practitioners, Doctoral Consortium and Blue Sky* (pp. 428–434). Springer Nature Switzerland. https://doi.org/10.1007/978-3-031-36336-8_67

Benedetto, L., Aradelli, G., Donvito, A., Lucchetti, A., Cappelli, A., & Buttery, P. (2024). Using LLMs to simulate students' responses to exam questions. In Y. Al-Onaizan, M. Bansal, & Y.-N. Chen (Eds.), *Findings of the Association for Computational Linguistics: EMNLP 2024* (pp. 11351–11368). Association for Computational Linguistics. https://doi.org/10.18653/v1/2024.findings-emnlp.663

Benedetto, L., Cappelli, A., Turrin, R., & Cremonesi, P. (2020). Introducing a Framework to Assess Newly Created Questions with Natural Language Processing. In I. I. Bittencourt, M. Cukurova, K. Muldner, R. Luckin, & E. Millán (Eds.), *Artificial Intelligence in Education* (pp. 43–54). Springer International Publishing. https://doi.org/10.1007/978-3-030-52237-7_4

Bulut, O., Gorgun, G., & Tan, B. (2024). Item Difficulty and Response Time Prediction with Large Language Models: An Empirical Analysis of USMLE Items. In E. Kochmar, M. Bexte, J. Burstein, A. Horbach, R. Laarmann-Quante, A. Tack, V. Yaneva, & Z. Yuan (Eds.), *Proceedings of the 19th Workshop on Innovative Use of NLP for Building Educational Applications (BEA 2024)* (pp. 522–527). Association for Computational Linguistics. https://aclanthology.org/2024.bea-1.44/

Clark, K., Luong, M.-T., Le, Q. V., & Manning, C. D. (2020). *ELECTRA: Pre-training Text Encoders as Discriminators Rather Than Generators* (arXiv:2003.10555). arXiv. https://doi.org/10.48550/arXiv.2003.10555

Devlin, J., Chang, M.-W., Lee, K., & Toutanova, K. (2019). BERT: Pre-training of Deep Bidirectional Transformers for Language Understanding. In J. Burstein, C. Doran, & T. Solorio (Eds.), *Proceedings of the 2019 Conference of the North American Chapter of the Association for Computational Linguistics: Human Language Technologies, Volume 1 (Long and Short Papers)* (pp. 4171–4186). Association for Computational Linguistics. https://doi.org/10.18653/v1/N19-1423

Drum, P. A., Calfee, R. C., & Cook, L. K. (1981). The Effects of Surface Structure Variables on Performance in Reading Comprehension Tests. *Reading Research Quarterly*,

*16*(4), 486–514. https://doi.org/10.2307/747313

Ferrara, S., Steedle, J. T., & Frantz, R. S. (2022a). Response Demands of Reading Comprehension Test Items: A Review of Item Difficulty Modeling Studies. *Applied Measurement in Education*, *35*(3), 237–253. https://doi.org/10.1080/08957347.2022.2103135

Ferrara, S., Steedle, J. T., & Frantz, R. S. (2022b). Response Demands of Reading Comprehension Test Items: A Review of Item Difficulty Modeling Studies. *Applied Measurement in Education*, *35*(3), 237–253. https://doi.org/10.1080/08957347.2022.2103135

Ha, L. A., Yaneva, V., Baldwin, P., & Mee, J. (2019). Predicting the Difficulty of Multiple Choice Questions in a High-stakes Medical Exam. In H. Yannakoudakis, E. Kochmar, C. Leacock, N. Madnani, I. Pilán, & T. Zesch (Eds.), *Proceedings of the Fourteenth Workshop on Innovative Use of NLP for Building Educational Applications* (pp. 11–20). Association for Computational Linguistics. https://doi.org/10.18653/v1/W19-4402

He, P., Liu, X., Gao, J., & Chen, W. (2021). *DeBERTa: Decoding-enhanced BERT with Disentangled Attention* (arXiv:2006.03654). arXiv. https://doi.org/10.48550/arXiv.2006.03654

Hsu, F.-Y., Lee, H.-M., Chang, T.-H., & Sung, Y.-T. (2018). Automated estimation of item difficulty for multiple-choice tests: An application of word embedding techniques. *Information Processing & Management*, *54*(6), 969–984. https://doi.org/10.1016/j.ipm.2018.06.007

Jiang, Z.-H., Yu, W., Zhou, D., Chen, Y., Feng, J., & Yan, S. (2020). ConvBERT: Improving BERT with Span-based Dynamic Convolution. *Advances in Neural Information Processing Systems*, *33*, 12837–12848. https://proceedings.neurips.cc/paper_files/paper/2020/hash/96da2f590cd7246bbde0051047b0d6f7-Abstract.html

Lane, S., Raymond, M. R., & Haladyna, T. M. (2016). *Handbook of test development* (Vol. 2). Routledge New York, NY. https://api.taylorfrancis.com/content/books/mono/download?identifierName=doi&identifierValue=10.4324/9780203102961&type=googlepdf

Li, M., Chen, H., Xiao, Y., Chen, J., Jiao, H., & Zhou, T. (2025). *Can LLMs Estimate Student Struggles? Human-AI Difficulty Alignment with Proficiency Simulation for Item Difficulty Prediction*. arXiv. https://doi.org/10.48550/arXiv.2512.18880

Li, M., Jiao, H., Zhou, T., Zhang, N., Peters, S., & Lissitz, R. W. (2025). Item Difficulty Modeling Using Fine-tuned Small and Large Language Models. *Educational and Psychological Measurement*, *85*(6), 1065–1090. https://doi.org/10.1177/00131644251344973

Liu, Y., Bhandari, S., & Pardos, Z. A. (2025). Leveraging LLM Respondents for Item Evaluation: A Psychometric Analysis. *British Journal of Educational Technology*, *56*(3), 1028–1052. https://doi.org/10.1111/bjet.13570

Liu, Y., Ott, M., Goyal, N., Du, J., Joshi, M., Chen, D., Levy, O., Lewis, M., Zettlemoyer, L., & Stoyanov, V. (2019). *RoBERTa: A Robustly Optimized BERT Pretraining Approach* (arXiv:1907.11692). arXiv. https://doi.org/10.48550/arXiv.1907.11692

Park, J.-W., Park, S.-J., Won, H.-S., & Kim, K.-M. (2024). Large Language Models are Students at Various Levels: Zero-shot Question Difficulty Estimation. In Y. Al-Onaizan, M. Bansal, & Y.-N. Chen (Eds.), *Findings of the Association for Computational Linguistics: EMNLP 2024* (pp. 8157–8177). Association for Computational Linguistics. https://doi.org/10.18653/v1/2024.findings-emnlp.477

Perkins, K., Gupta, L., & Tammana, R. (1995). Predicting item difficulty in a reading comprehension test with an artificial neural network. *Language Testing*, *12*(1), 34–53. https://doi.org/10.1177/026553229501200103

Peters, S., Zhang, N., Jiao, H., Li, M., & Zhou, T. (2025). Review of Text-Based Approaches to Item Difficulty Modeling in Large-Scale Assessments. In J. Wilson, C. Ormerod, & M. Beiting Parrish (Eds.), *Proceedings of the Artificial Intelligence in Measurement and Education Conference (AIME-Con): Coordinated Session Papers* (pp. 37–47). National Council on Measurement in Education (NCME). https://aclanthology.org/2025.aimecon-sessions.4/

Sokolova, M., & Lapalme, G. (2009). A systematic analysis of performance measures for classification tasks. *Information Processing & Management*, *45*(4), 427–437. https://doi.org/10.1016/j.ipm.2009.03.002

Štěpánek, L., Dlouhá, J., & Martinková, P. (2023). Item Difficulty Prediction Using Item Text Features: Comparison of Predictive Performance across Machine-Learning Algorithms. *Mathematics*, *11*(19), 4104. https://doi.org/10.3390/math11194104

Wang, L., Yang, N., Huang, X., Yang, L., Majumder, R., & Wei, F. (2024, February 8). *Multilingual E5 Text Embeddings: A Technical Report*. arXiv.Org.

https://arxiv.org/abs/2402.05672v1

Wauters, K., Desmet, P., & Van Den Noortgate, W. (2012). Item difficulty estimation: An auspicious collaboration between data and judgment. *Computers & Education*, *58*(4), 1183–1193. https://doi.org/10.1016/j.compedu.2011.11.020

Yancey, K. P., Runge, A., LaFlair, G., & Mulcaire, P. (2024). BERT-IRT: Accelerating Item Piloting with BERT Embeddings and Explainable IRT Models. In E. Kochmar, M. Bexte, J. Burstein, A. Horbach, R. Laarmann-Quante, A. Tack, V. Yaneva, & Z. Yuan (Eds.), *Proceedings of the 19th Workshop on Innovative Use of NLP for Building Educational Applications (BEA 2024)* (pp. 428–438). Association for Computational Linguistics. https://aclanthology.org/2024.bea-1.35/

Yi, X., Sun, J., & Wu, X. (2024). Novel Feature-Based Difficulty Prediction Method for Mathematics Items Using XGBoost-Based SHAP Model. *Mathematics*, *12*(10), 1455. https://doi.org/10.3390/math12101455

Zhou, Y., & Tao, C. (2020). *Multi-task BERT for problem difficulty prediction*. 213–216. https://doi.org/10.1109/CISCE50729.2020.00048

Zotos, L., Rijn, H. van, & Nissim, M. (2025). *Are You Doubtful? Oh, It Might Be Difficult Then! Exploring the Use of Model Uncertainty for Question Difficulty Estimation* (arXiv:2412.11831). arXiv. https://doi.org/10.48550/arXiv.2412.11831

## Appendix A

### Zero-shot Prompt

Classify the difficulty of the provided Reading and Writing item as Easy, Medium, or Hard.

Please output only one word: Easy, Medium, or Hard.

New item: The following text is from H.D.'s 1916 poem "Mid-Day." In the poem, the speaker is on a path in an outdoor setting. A slight wind shakes the seed-pods — my thoughts are spent as the black seeds… Choice A is incorrect because the text does not indicate that the seedpods are the cause of the speaker's state of mind…

Predicted difficulty:

## Appendix B

### Few-shot Prompt

Classify the difficulty of the provided Reading and Writing item as Easy, Medium, or Hard.

Please output only one word: Easy, Medium, or Hard.

The following are 3 examples from the same skill but with different difficulty levels.

Please analyze the reasons why they are classified at different levels, and then classify the difficulty of the new item.

Example 1: The following text is adapted from Pam Muñoz Ryan's 2020 novel Mañanaland . ……

Difficulty 1: Easy

Example 2: Studying late nineteenth- and early twentieth-century artifacts from an agricultural and domestic site in Texas, archaeologist Ayana O. Flewellen found that Black women employed as farm workers utilized hook-and-eye closures to fasten their clothes at the waist, giving themselves a silhouette similar to the one that was popular in contemporary fashion and typically achieved through more restrictive garments such as corsets. ……

Difficulty 2: Hard

Example 3: The Bayeux Tapestry, from eleventh-century France, depicts 75 scenes over 250 feet of fabric. …….

Difficulty 3: Medium

New item:

The following text is from H.D.'s 1916 poem "Mid-Day." In the poem, the speaker is on a path in an outdoor setting. ……

Predicted difficulty:

## Appendix C

## Persona Prompt

Suppose you are a student taking the SAT reading exam. You are a weak/average/good student with low/medium/high-level English proficiency.

Analyze the difficulty levels of the question. The difficulty levels contain 3 categories: easy, medium, and hard. Analyze the difficulty step by step, and provide the final category in \boxed {…}:

Item: